\documentclass[10pt,twocolumn,letterpaper]{article}
\usepackage{multirow}
\usepackage{cvpr}
\usepackage{times}
\usepackage{epsfig}
\usepackage{graphicx}
\usepackage{amsmath}
\usepackage{amssymb}
\usepackage{subfiles}
\usepackage{float}
\usepackage{cvpr}
\usepackage{times}
\usepackage{setspace}
\usepackage{pifont}

\usepackage[stable]{footmisc}

\cvprfinalcopy 

\ifcvprfinal\fi
\begin{document}

\title{Towards Photo-Realistic Virtual Try-On by Adaptively\\  Generating$\leftrightarrow$Preserving Image Content}

\author{
Han Yang\textsuperscript{1,2}\footnotemark\,\,\,\,\, 
Ruimao Zhang\textsuperscript{2}\,\,\,\,\,
Xiaobao Guo\textsuperscript{2}\,\,\,\,\,
Wei Liu\textsuperscript{3}\,\,\,\,\,
Wangmeng Zuo\textsuperscript{1}\,\,\,\,\,
Ping Luo\textsuperscript{4}\\
\centerline{
\textsuperscript{1}Harbin Institute of Technology\,\,\,\,\,
\textsuperscript{2}SenseTime Research}\\
\centerline{
\textsuperscript{3}Tencent AI Lab\,\,\,\,\,
\textsuperscript{4}The University of HongKong}
\\
{\tt\small \{yanghancv, wmzuo\}@hit.edu.cn, wl2223@columbia.edu, pluo@cs.hku.hk}\\
{\tt\small \{zhangruimao, guoxiaobao\}@sensetime.com}
}

\begin{figure}[htb]
\twocolumn[{
\renewcommand\twocolumn[1][]{#1}%
\maketitle
\vspace{-30pt}
\begin{center}
  \centering
  \includegraphics[width=1\textwidth]{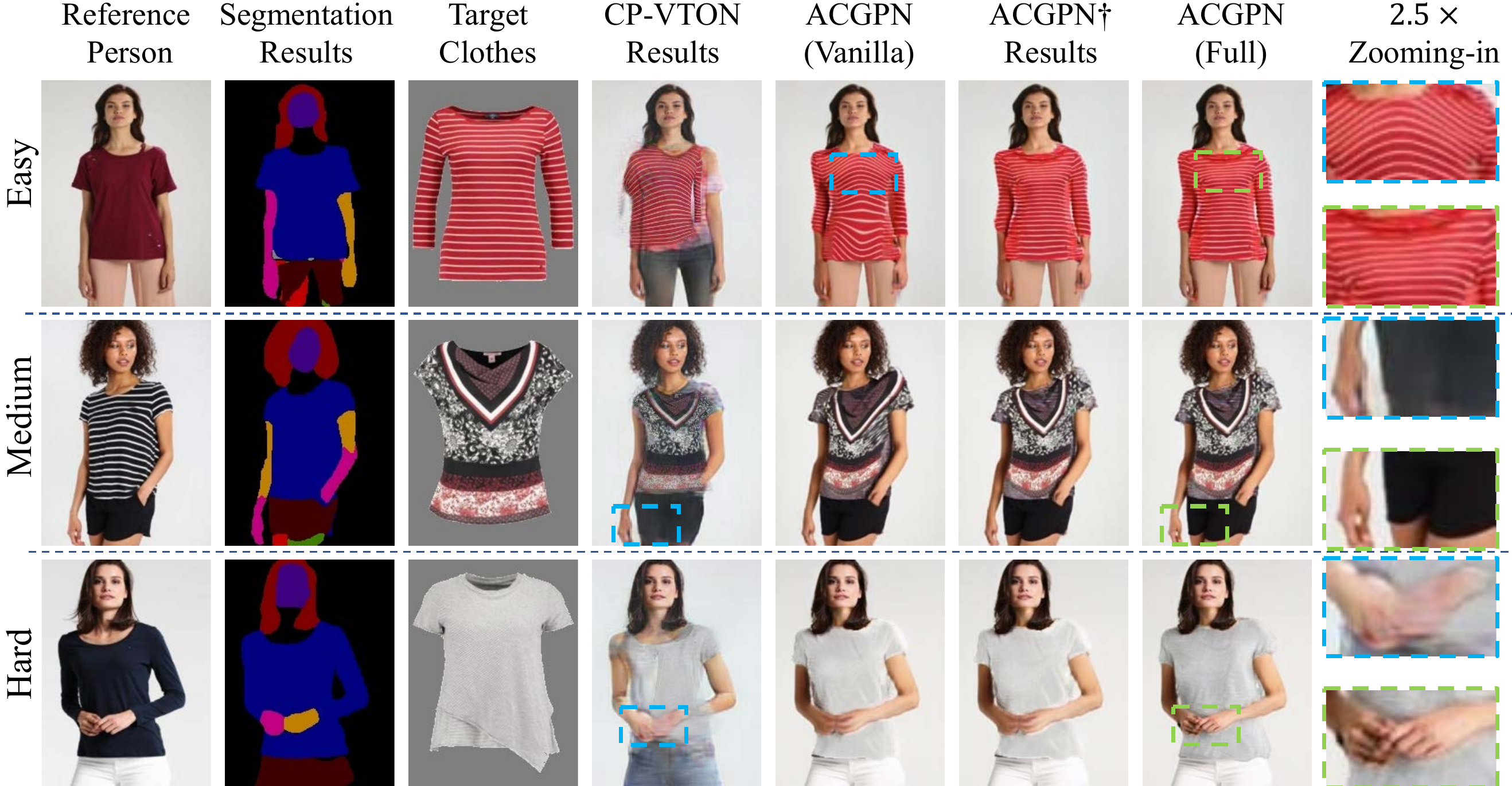}
  \vspace{-5pt}
\end{center}
\caption{\footnotesize We define the difficulty level of try-on task to easy, medium, and hard based on current works. Given a target clothing image and a reference image, our method synthesizes a person in target clothes while preserving photo-realistic details such as character of clothes (texture, logo), posture of person (non-target body parts, bottom clothes), and identity of person. ACGPN (Vanilla) indicates ACGPN without warping constraint or non-target body composition, ACGPN$\dag$ adds warping constraint on ACGPN (Vanilla). Also, zooming-in of the greatly improved regions are given on the right.}
\label{fig:firstpage}
\vspace{25pt}
}
]
\end{figure}


 \vspace{-35pt}
\begin{abstract}
 \vspace{-10pt}

\thispagestyle{empty}

\footnotetext{This work was done when Han Yang was a Research Intern at SenseTime Research
.}

Image visual try-on aims at transferring a target clothing  image onto a reference person, and has become a hot  topic in recent years. Prior arts usually focus on preserving the character of a clothing  image (\eg texture, logo, embroidery)
when warping it to arbitrary human pose. However, it remains a big challenge to generate photo-realistic try-on images
when large occlusions and human poses are presented in
the reference person (Fig.~\ref{fig:firstpage}). To address this issue, we propose a novel
visual try-on network, namely Adaptive Content Generating and Preserving Network (ACGPN).
In particular, ACGPN first
predicts semantic layout of the reference image that will be changed after try-on (\eg long sleeve shirt$\rightarrow$arm, arm$\rightarrow$jacket), and then determines whether its image content needs to be generated
or preserved according to the predicted semantic layout,
leading to photo-realistic try-on and rich clothing details. ACGPN generally involves
three major modules.
First, a semantic layout generation module utilizes semantic segmentation of the reference image to progressively predict the desired semantic layout after try-on.
%
Second, a
clothes warping module warps clothing  images
according to the generated semantic layout, where a second-order difference constraint is introduced to stabilize the warping process during training.
Third, an inpainting module for content fusion integrates all information (\eg reference image, semantic layout, warped clothes) to adaptively produce each semantic part of human body.
In comparison to the state-of-the-art methods,
ACGPN can generate photo-realistic images with much better perceptual quality and richer fine-details.

\end{abstract}

\section{Introduction}

Motivated by the rapid development of image synthesis~\cite{DBLP:conf/cvpr/IsolaZZE17,DBLP:conf/cvpr/Park0WZ19,DBLP:conf/iclr/KarrasALL18,DBLP:conf/cvpr/KarrasLA19}
, image-based visual try-on~\cite{DBLP:conf/iccvw/JetchevB17,DBLP:conf/cvpr/HanWWYD18} aiming to transfer the target clothing item onto reference person has achieved much attention in recent years.
Although considerable progress has been made~\cite{DBLP:conf/eccv/WangZLCLY18,DBLP:journals/corr/abs-1902-11026,Yu_2019_ICCV,DBLP:conf/intcompsymp/ChenTC18}, it remains a challenging task to build up the photo-realistic  virtual try-on system for real-world scenario,
partially ascribing to the semantic and geometric difference between the target clothes and reference images, as well as the interaction occlusions between the torso and limbs.

To illustrate the limitations of existing visual try-on methods,
we divide the VITON dataset~\cite{DBLP:conf/cvpr/HanWWYD18} into three subsets of difficulty levels according to the human pose in 2D reference images.
%
%
As shown in Fig.~\ref{fig:firstpage}, the first row gives an easy sample from VITON dataset~\cite{DBLP:conf/cvpr/HanWWYD18}, where the person in the image is represented with a standard posture, \ie face forward and hands down.
In such case, the methods only need to align the semantic regions between the reference and target images.
Some pioneering synthesized-based methods~\cite{DBLP:conf/iccvw/JetchevB17,DBLP:conf/intcompsymp/ChenTC18,DBLP:conf/eccv/RajSCHCL18,DBLP:conf/cvpr/ChoiCKH0C18,DBLP:journals/corr/abs-1906-07251} belong to this category.
From the second row, the image with medium-level difficulty is generally with torso posture changes.
And several models~\cite{DBLP:conf/cvpr/HanWWYD18,DBLP:conf/eccv/WangZLCLY18,DBLP:journals/corr/abs-1902-11026,Yu_2019_ICCV} have been suggested to preserve the character of the clothes, such as texture, logo, embroidery and so on.
Such goal is usually attained by developing advanced warping algorithms to match the reference image with clothes deformation.
The last row of Fig.~\ref{fig:firstpage} presents a hard example,
where postural changes occur on both the torso and the limbs, leading to the spatial interactions between the clothing regions and human body parts, \eg occlusions, disturbances, and deformation.
Therefore, the algorithm is required to understand the spatial layout of the foreground and background objects in the reference image,
and adaptively preserve such occlusion relationship in the try-on process.
However, content generation and preservation remain an uninvestigated issue in virtual try-on.

\begin{figure*}[t]
\begin{center}
\includegraphics[width=0.85\linewidth]{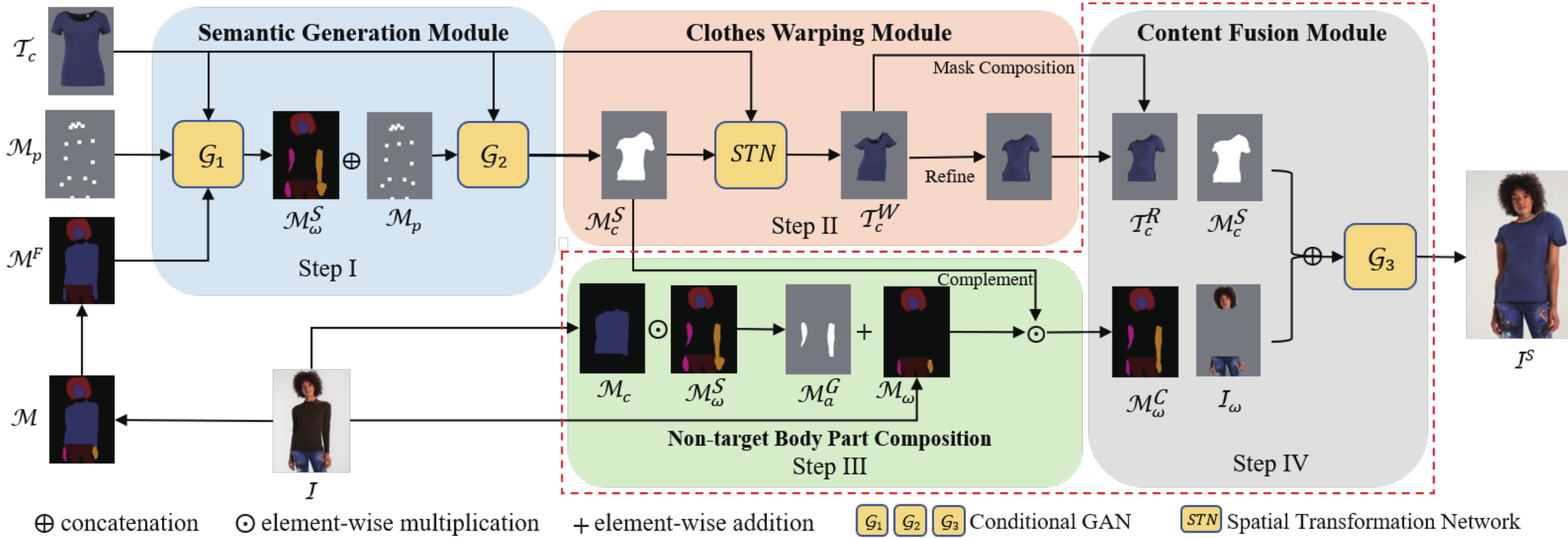}
\vspace{-10pt}
\end{center}
   \caption{\footnotesize The overall architecture of our ACGPN. (1) In Step I, the Semantic Generation Module (SGM) takes target clothing  image $\mathcal{T}_c$, the pose map $\mathcal{M}_{p}$, and the fused body part mask $\mathcal{M}^{F}$ as the input to predict the semantic layout and to output synthesized body part mask $\mathcal{M}^{S}_{\omega}$ and target clothing mask $\mathcal{M}^{S}_{c}$; (2) In Step II, the Clothes Warping Module (CWM) warps the target clothing  image to $\mathcal{T}^{R}_{c}$ according to the predicted semantic layout, where a second-order difference constraint is introduced to stabilize the warping process; (3) In Steps III and IV, the Content Fusion Module (CFM) first produces  the composited body part mask $\mathcal{M}^{C}_{\omega}$ using the original clothing mask $\mathcal{M}_{c}$, the synthesized clothing mask $\mathcal{M}^{S}_{c}$, the body part mask $\mathcal{M}_{\omega}$, and the synthesized body part mask $\mathcal{M}^{S}_{\omega}$ and then exploits a fusion network to generate the try-on images $\mathcal{I}^{S}$ by utilizing the information $\mathcal{T}^{R}_{c}$, $\mathcal{M}^{S}_{c}$, and body part image $\mathcal{I}_{\omega}$ from previous steps.}
\label{fig:2}
\vspace{-10pt}
\end{figure*}

To address the above limitations, this paper presents a novel Adaptive Content Generation and Preservation Network (ACGPN),
which first predicts the semantic layout of the reference image and then adaptively determines the content generation or preservation according to the predicted semantic layout.
Specially, the ACGPN consists of three major modules as shown in Fig.~\ref{fig:2}.
The first one is the Semantic Generation Module (SGM), which uses the semantic segmentation of body parts and clothes to progressively generate the mask of exposed body parts (\ie synthesized body part mask) and the mask of warped clothing regions.
As opposed to prior arts, the proposed SGM generate semantic masks in a two-stage fashion to generate the body parts first and synthesize clothing mask progressively, which makes the original clothes shape in reference image completely agnostic to the network.
The second part is the Clothes Warping Module (CWM), which is designed to warp clothes according to the generated semantic layout.
Going beyond the Thin-Plate Spline based methods~\cite{DBLP:conf/cvpr/HanWWYD18,DBLP:conf/eccv/WangZLCLY18,DBLP:journals/corr/abs-1902-11026}
a second-order difference constraint is also introduced to the Warping loss to make the warping process more stable,
especially for the clothes with the complex texture.
Finally, the Content Fusion Module (CFM) integrates the information from the synthesized body part mask, the warped clothing  image, and the original body part image to adaptively determine the generation or preservation of the distinct human parts in the synthesized image.

With the above modules, ACGPN adopts a split-transform-merge strategy to generate a spatial configuration aware try-on image.
Experiments on the VITON dataset~\cite{DBLP:conf/eccv/WangZLCLY18} show that our ACGPN not only promotes the visual quality of generated images for the easy and medium difficulty levels (see  Fig.~\ref{fig:firstpage}),
but also is effective in handling the hard try-on case with the semantic region intersections in an elegant way and produces photo-realistic results.

The main contributions of this paper can be summarized as follows.
(1) We propose a new image-based virtual try-on network, \ie ACGPN, which greatly improves the try-on quality in semantic alignment, character retention and layout adaptation.
(2) We for the first time take the semantic layout into consideration to generate the photo-realistic try-on results. A novel adaptive content generation and preservation scheme is proposed.
(3) A novel second-order difference constraint makes the training process of warping module more stable, and improves the ability of our method to handle complex textures on clothes.
(4) Experiments demonstrate that the proposed method can generate photo-realistic images that outperform the state-of-the-art methods both qualitatively and quantitatively.

\vspace{-5pt}
\section{Related Work}
\textbf{Generative Adversarial Networks}. Generative Adversarial Networks (GAN) has greatly facilitated the improvements and advancements in image synthesis~\cite{DBLP:conf/cvpr/IsolaZZE17,DBLP:conf/cvpr/Park0WZ19,DBLP:conf/iclr/KarrasALL18,DBLP:conf/cvpr/KarrasLA19} and manipulation~\cite{DBLP:journals/corr/abs-1902-06838,lee2019maskgan,DBLP:journals/corr/abs-1906-00884}. GAN generally consists of a generator and a discriminator. The generator learns to generate realistic images to deceive the discriminator, while the discriminator learns to distinguish the synthesized images from the real ones.
Benefited from the powerful ability of GAN, it enjoys pervasive applications on tasks such as style transfer~\cite{Zhu_2017_ICCV,DBLP:conf/cvpr/ChoiCKH0C18}, image inpainting~\cite{DBLP:conf/cvpr/XiongYLYLBL19,DBLP:journals/tog/IizukaS017,DBLP:journals/corr/abs-1806-03589,DBLP:conf/cvpr/Yu0YSLH18,liu2018image}, and image editing~\cite{DBLP:journals/corr/abs-1902-06838,DBLP:journals/corr/abs-1906-00884,lee2019maskgan,DBLP:conf/cvpr/Park0WZ19}. The extensive applications of GAN further demonstrate its superiority in image synthesis.

\textbf{Fashion Analysis and Synthesis}. Fashion related tasks recently have received considerable attention due to their great potential in real-world applications. Most of the existing works focus on clothing compatibility and matching learning~\cite{DBLP:journals/tmm/LiCZL17,DBLP:conf/ijcai/IwataWS11,DBLP:conf/iccv/VeitKBMBB15}, clothing landmark detection~\cite{liu2016fashion,yan2017unconstrained,ge2019deepfashion2,lee2019global}, and fashion image analysis~\cite{hsiao2019fashion++,han2019finet,liu2018deep}. Virtual try-on is among the most challenging tasks in fashion analysis.

\textbf{Virtual Try-on}. Virtual try-on has been an attractive topic  even before the renaissance of deep learning~\cite{zhou2012image,DBLP:conf/ismar/EharaS06,tanaka2009texture,DBLP:journals/tvcg/HauswiesnerSR13}. In the recent years, along with the progress in deep neural networks, virtual try-on has raised more and more interest due to its great potential in many real applications. Existing deep learning based methods on virtual try-on can be classified as 3D model based approaches~\cite{DBLP:journals/cgf/SantestebanOC19,DBLP:journals/tog/BrouetSBC12,DBLP:journals/tog/GuanRHWB12,DBLP:journals/tog/Pons-MollPHB17,DBLP:journals/tog/RohmerPCHS10} and 2D image based ones~\cite{DBLP:conf/cvpr/HanWWYD18,DBLP:conf/eccv/WangZLCLY18,DBLP:journals/corr/abs-1902-11026,DBLP:conf/iccvw/JetchevB17}, where the latter can be further categorized based on whether to keep the posture or not. Dong et al.~\cite{DBLP:journals/corr/abs-1902-11026} presents a multi-pose guided image based virtual try-on network. Analogous to our ACGPN, most existing try-on methods focus on the task of keeping the posture and identity.
Methods such as VITON~\cite{DBLP:conf/cvpr/HanWWYD18}, CP-VTON~\cite{DBLP:conf/eccv/WangZLCLY18} use coarse human shape and pose map as the input to generate a clothed person.
While methods such as SwapGAN~\cite{DBLP:journals/tmm/LiuCLL19}, SwapNet~\cite{DBLP:conf/eccv/RajSCHCL18} and VTNFP~\cite{Yu_2019_ICCV} adopt semantic segmentation~\cite{DBLP:journals/pr/ZhangYPWWL19} as input to synthesize clothed person.
Table ~\ref{tab:Overview} presents an overview of several representative methods.
VITON~\cite{DBLP:conf/cvpr/HanWWYD18} exploits a Thin-Plate Spline (TPS)~\cite{duchon1977splines}  based warping method to first deform the inshop clothes and map the texture to the refined result with a composition mask.
CP-VTON~\cite{DBLP:conf/eccv/WangZLCLY18} adopts a similar structure of VITON but uses a neural network to learn the transformation parameters of TPS warping rather than using image descriptors, and achieves more accurate alignment results.
CP-VTON and VITON only focus on the clothes, leading to coarse and blurry bottom clothes and posture details. VTNFP~\cite{Yu_2019_ICCV} alleviates this issue by simply concatenating the high-level features extracted from body parts and bottom clothes, thereby generating better results than CP-VTON and VITON.
However, blurry body parts and artifacts still remain abundantly in the results because it ignores the semantic layout of the reference image. 

\begin{table}
\setlength{\baselineskip}{2em} 
\renewcommand\tabcolsep{1pt} %
\small
\begin{center}

\begin{tabular}{llccccc}
\hline
&&CA~\cite{DBLP:conf/iccvw/JetchevB17}&VI~\cite{DBLP:conf/cvpr/HanWWYD18}&CP~\cite{DBLP:conf/eccv/WangZLCLY18}&VT~\cite{Yu_2019_ICCV}&Ours\\
\hline\hline
\parbox[t]{2mm}{\multirow{3}{*}{\rotatebox[origin=c]{90}{\tiny Representation}}}&Use Coarse Shape& $\times$  &$\surd$&$\surd$&$\surd$&$\times$\\
&Use Pose& $\times$ &$\surd$&$\surd$&$\surd$&$\surd$\\
&Use Segmentation& $\times$ &$\times$&$\times$&$\surd$&$\surd$\\
\hline

\parbox[t]{2mm}{\multirow{3}{*}{\rotatebox[origin=c]{90}{\tiny Preservation}}}&Texture& $\times$ &$\surd$&$\surd$&$\surd$&$\surd$\\
&Non-target clothes& $\times$ &$\times$&$\times$&$\surd$&$\surd$\\
&Body Parts& $\times$ &$\times$&$\times$&$\times$&$\surd$\\
\hline

\parbox[t]{2mm}{\multirow{3}{*}{\rotatebox[origin=c]{90}{\tiny Problem}}}&Semantic Alignment& $\surd$ &$\surd$&$\surd$&$\surd$&$\surd$\\
&Character Retention& $\times$ &$\surd$&$\surd$&$\surd$&$\surd$\\
&Layout Adaptation& $\times$ &$\times$&$\times$&$\times$&$\surd$\\

\hline
\end{tabular}
\vspace{10pt}

\caption{\footnotesize Comparison of representative virtual try-on methods. CA refers to CAGAN~\cite{DBLP:conf/iccvw/JetchevB17}; VI refers to VITON~\cite{DBLP:conf/cvpr/HanWWYD18}; CP refers to CP-VTON~\cite{DBLP:conf/eccv/WangZLCLY18}, and VT refers to VTNFP~\cite{Yu_2019_ICCV}. We compare ACGPN with four popular image-based virtual try-on methods, \ie CAGAN, VITON, CP-VTON and VTNFP, and we compare them from four aspects: representations as input, preservation of source information, and problems to solve.}
\label{tab:Overview}

\vspace{-20pt}

\end{center}
\end{table}

In Table \ref{tab:Overview}, CAGAN uses analogy learning to transfer the garment onto reference person, but can only preserve the color and coarse shape. VITON presents a coarse-to-fine structure which utilizes coarse shape and pose map to ensure generalization on arbitrary clothes. CP-VTON adopts the same pipeline of VITON, while changing the warping module into a learnable network. These two methods perform quite well with retention of the character of clothes, but overlook the non-target body parts and bottom clothes. VTNFP ameliorates this ignorance by adding weak supervision of original body parts as well as bottom clothes to help preserve more details, which generates more realistic images than CAGAN, VITON and CP-VTON; however, VTNFP results still have a large gap between photo-realistic due to their artifacts.


\section{Adaptive Content Generating and Preserving Network}
The proposed ACGPN is composed of three modules, as shown in Fig.~2. First, the Semantic Generation Module (SGM)  progressively generates the mask of the body parts and the mask of the warped clothing regions via semantic segmentation, yielding semantic alignment of the spatial layout.
Second, the Clothes Warping Module (CWM) is designed to warp target clothing  image according to the warped clothing mask, where we introduce a second-order difference constraint on Thin-Plate Spline (TPS)~\cite{duchon1977splines}  to produce geometric matching yet character retentive clothing  images.
Finally, Steps 3 and 4 are united in the Content Fusion Module (CFM), which integrates the information from previous modules to adaptively determine the generation or preservation of the distinct human parts in output synthesized image. Non-target body part composition is able to handle different scenarios flexibly in try-on task while mask inpainting fully exploits the layout adaptation ability of the ACGPN when dealing with the images from easy, medium, and hard levels of difficulty.

\subsection{Semantic Generation Module (SGM)}

The semantic generation module (SGM) is proposed to separate the target clothing region as well as to preserve the body parts (\ie arms) of the person, without changing the pose and the rest human body details.
Many previous works focus on the target clothes but overlook human body generation by only feeding the coarse body shape directly into the network, leading to the loss of the body part details. To address this issue, the mask generation mechanism is adopted in this module to generate semantic segmentation of body parts and target clothing region precisely.

Specifically, given a reference image $\mathcal{I}$, and its corresponding mask $\mathcal{M}$, arms $\mathcal{M}_{a}$ and torso $\mathcal{M}_{t}$ are first fused into an indistinguishable area, resulting in the fused map $\mathcal{M}^{F}$ shown in Fig.~\ref{fig:2} as one of the inputs to SGM.
Following a two-stage strategy, the try-on mask generation module first synthesize the masks of the body parts $\mathcal{M}^{S}_{\omega}$ ($\omega = \{h, a, b\}$ (h:head, a:arms, b:bottom clothes)), which helps to adaptively preserve body parts instead of coarse feature in the subsequent steps.
As shown in Fig.~\ref{fig:2}, we train a body parsing GAN $\mathcal{G}_{1}$ to generate $\mathcal{M}^{S}_{\omega}$ by leveraging the information from the fused map $\mathcal{M}^{F}$, the pose map $\mathcal{M}_{p}$, and the target clothing  image $\mathcal{T}_{c}$. Using the generated information of body parts, its corresponding pose map and target clothing  image, it is tractable to get the estimated clothing region. In the second stage, $\mathcal{M}^{S}_{\omega}$, $\mathcal{M}_{p}$ and $\mathcal{T}_{c}$ are combined to generate the synthesized mask of the clothes $\mathcal{M}^{S}_{c}$ by $\mathcal{G}_{2}$.


For training SGM, both stages adopt the conditional generative adversarial network (cGAN), in which a U-Net structure is used as the generator while a discriminator given in pix2pixHD~\cite{DBLP:conf/cvpr/Wang0ZTKC18} is deployed to distinguish generated masks from their ground-truth masks.
For each of the stages, the CGAN loss can be formulated as
\begin{equation}
\begin{split}
    \mathcal{L}_{1}=&\mathbb{E}_{x,y}\left[\log\left(\mathcal{D}\left(x,y\right)\right)\right]\\
    &+\mathbb{E}_{x,z}\left[\log\left(1-\mathcal{D}\left(x,\mathcal{G}\left(x,z\right)\right)\right)\right],
\end{split}
\label{GANLOSS}
\end{equation}
where $x$ indicates the input and $y$ is the ground-truth mask. $z$ is the noise which is an additional channel of input sampled from standard normal distribution.

The overall objective function for each stage of the proposed try-on mask generation module is formulated as $\mathcal{L}_{m}$,
\begin{equation}
    \mathcal{L}_{m}=\lambda_{1}  \mathcal{L}_{1}+\lambda_{2} \mathcal{L}_{2},
\label{module1loss}
\end{equation}
where $\mathcal{L}_{2}$ is the pixel-wise cross entropy loss~\cite{goodfellow2016deep}, which improves the quality of synthesized masks from generator with more accurate semantic segmentation results. $\lambda_{1}$ and $\lambda_{2}$ are the trade-off parameters for each loss term in Eq.~(\ref{module1loss}), which are set to 1 and 10, respectively in our experiments.

%
%
%
%

The two-stage SGM can serve as a core component for accurate understanding of body-parts and clothes layouts in visual try-on and guiding
the adaptive preserving of image content by composition. We also believe SGM is effective for other tasks that need to partition semantic layout.

\subsection{Clothes Warping Module (CWM)}

\begin{figure}[t]
\begin{center}
\includegraphics[width=0.85\linewidth]{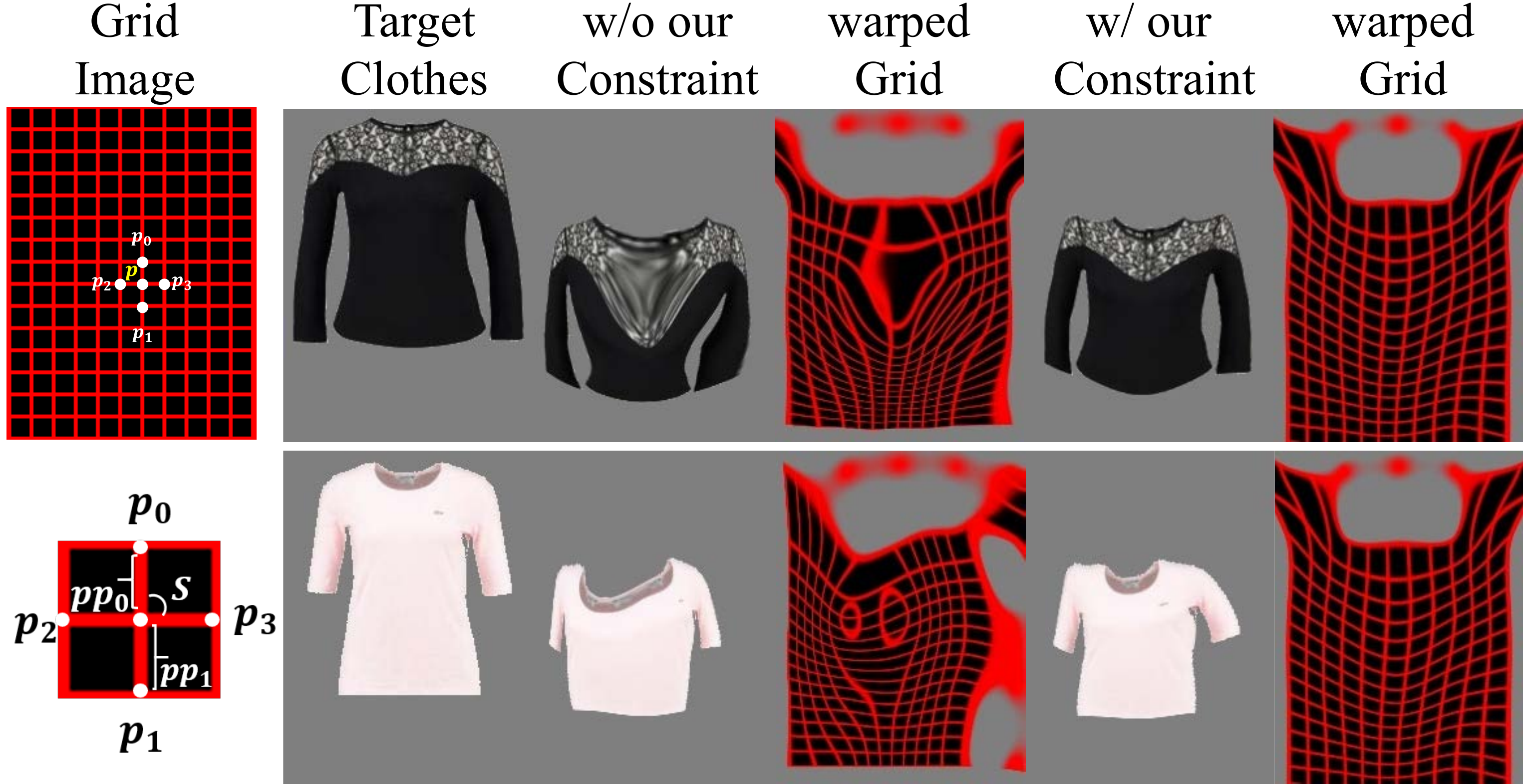}
\vspace{-10pt}
\end{center}
   \caption{\footnotesize Comparison of STN warping results with and without the second-order difference  constraint.}
\label{fig:warpexample}
\vspace{-10pt}
\end{figure}

Clothes warping aims to fit the clothes into the shape of target clothing region with visually natural deformation according to human pose as well as to retain the character of the clothes.
However, simply training a Spatial Transformation Network (STN)~\cite{jaderberg2015spatial} and applying Thin-Plate Spline (TPS)~\cite{duchon1977splines}  cannot ensure the precise transformation especially when dealing with hard cases (\ie the clothes with complex texture and rich colors), leading to misalignment and blurry results. To address these problems, we introduce a second-order difference constraint on the clothes warping network to realize geometric matching and character retention. As shown in Fig.~\ref{fig:warpexample}, compared to the result with our proposed constraint, target clothes transformation without the constraint shows obvious distortion on shape and unreasonable mess on texture.

Formally, given $\mathcal{T}_{c}$ and $\mathcal{M}^{S}_{c}$ as the input, we train the STN to learn the mapping between them. The warped clothing  image $\mathcal{T}^{W}_{c}$ is transformed by the learned parameters from STN, where we introduce the following constraint $\mathcal{L}_{3}$ as a loss term,
\begin{equation}
 \begin{split}
   &\mathcal{L}_{3} \!=\!
   \sum_{p\in\mathbf{P}}\lambda_{r}\left| \lVert pp_0\rVert_2 \!-\! \lVert pp_1\rVert_2 \right| \!+\! \left| \lVert pp_2\rVert_2 \!-\! \lVert pp_3\rVert_2 \right|\\
   & \!+\! \lambda_{s} \left(\left| S\left(p,p_0\right) \!-\! S\left(p,p_1\right) \right| \!+\! \left| S\left(p,p_2\right) \!-\! S\left(p,p_3\right) \right| \right),
    \end{split}
 \end{equation}
 where $\lambda_{r}$ and $\lambda_{s}$ are the trade-off hyper-parameters. Practically we can minimize $max(\mathcal{L}_{3}\!-\!\Delta,0)$ for restriction, and $\Delta$ is a hyper-parameter. As illustrated in Fig.~\ref{fig:warpexample}, $p(x,y)$ represents a certain sampled control point and $p_0(x_0,y_0), p_1(x_1,y_1), p_2(x_2,y_2), p_3(x_3,y_3)$ are the top, bottom, left, right sampled control points of $p(x,y)$, respectively in the whole control points set $\mathbf{P}$; $S\left(p,p_i\right)=\frac{y_i-y}{x_i-x}$ ($i$ = {0, 1, 2, 3}) is the slope between two points.
 $\mathcal{L}_{3}$ is proposed to serve as a constraint on TPS transformation by minimizing the metric distance of two neighboring intervals for each axis and the distance of slopes, which
 maintains the collinearity,parallelism and immutability property of affine transformation. To avoid divided-by-zero error, the actual implementation of the second term is
 \begin{equation}
    \begin{split}
       &| S\left(p,p_i\right)\!-\! S\left(p,p_j\right)|\\
         &=|(y_i-y)(x_j-x)-(y_j-y)(x_i-x)|,
    \end{split}
 \end{equation}
where $(i,j)\in \{(0,1),(2,3)\}$.
The warping loss can be represented as $\mathcal{L}_{w}$, which measures the loss between the warped clothing  image $\mathcal{T}^{W}_{c}$ and its ground-truth $\mathcal{I}_{c}$,
 \begin{equation}
     \mathcal{L}_{w}=\mathcal{L}_{3} + \mathcal{L}_{4},
 \end{equation}
where $\mathcal{L}_{4} =\lVert \mathcal{T}^{W}_{c}-\mathcal{I}_{c}\rVert_1 $. The warped clothes are then fed into the refinement network to further generate more details, where a learned matrix $\mathbf{{\alpha}} (0 \leq {\alpha}_{ij} \leq 1)$ is then utilized to finally combine the two clothing  images as the refined clothing  image $\mathcal{T}^{R}_{c}$ by
\begin{equation}
    \mathcal{T}^{R}_{c}=(1-\alpha)\odot\mathcal{T}^{W}_{c}+\alpha \odot\mathcal{T}^{R}_{c},
\label{refineclothes}
\end{equation}
where $\odot$ denotes element-wise multiplication. $\mathbf{{\alpha}}$ is also restricted by an regularization term (refer to CP-VTON~\cite{DBLP:conf/eccv/WangZLCLY18}) and VGG loss is also introduced on $\mathcal{T}^{R}_{c}$ and $\mathcal{T}^{W}_{c}$. For better quality, GAN loss can be also used here.
Consequently, the refined clothing image can fully retain the character of the target clothes.
%
%
%
%
%
We believe our formulation of CWM
is effective in enforcing the collinearity of local affine transforms while maintaining the flexibility of TPS warping globally, which is beneficial to produce geometrically matched and realistic warped results.

\subsection{Content Fusion Module (CFM)}
Going beyond semantic alignment and character retention, it remains a great challenge to realize layout adaptation on visual try-on task. To this end, both the target clothing region is required to clearly rendered, and fine-scale details of body parts (\ie finger gaps) are needed to be adaptively preserved.
Existing methods usually adopt the coarse body shape as a cue to generate the final try-on images, and fail to reconstruct fine details.
In contrast, the proposed content fusion module (CFM) is composed of two main steps, \ie Steps 3 and 4 in Fig.~\ref{fig:2}. In particular, Step 3 is designed to fully maintain the untargeted body parts as well as adaptively preserve the changeable body part (\ie arms). Step 4 fills in the changeable body part by utilizing the masks and images generated from previous steps accordingly by an inpainting based fusion GAN, $\mathcal{G}_3$ in Fig.~\ref{fig:2}.

\textbf{Non-target Body Part Composition}. The composited body mask $\mathcal{M}^{C}_{\omega}$ is composed by original body part mask $\mathcal{M}_{\omega}$, the generated body mask $\mathcal{M}^{G}_{a}$ which is the region for generation, and synthesized clothing mask $\mathcal{M}^{S}_{c}$ according to 
\begin{equation}
    \mathcal{M}^{G}_{a}=\mathcal{M}^{S}_{\omega}\odot\mathcal{M}_{c},
\end{equation}
\begin{equation}
\label{euq:labelmask}
    \mathcal{M}^{C}_{\omega}=(\mathcal{M}^{G}_{a}+\mathcal{M}_{\omega})\odot(1-\mathcal{M}^{S}_{c}),
\end{equation}
\begin{equation}
\label{euq:mask}
    \mathcal{I}_{\omega}=\mathcal{I}_{\omega'}\odot(1-\mathcal{M}^{S}_{c}),
\end{equation}
where $\odot$ denotes element-wise multiplication, and Eq.~(\ref{euq:mask}) is not shown in Fig.~\ref{fig:2} for simplicity; $\mathcal{I}_{\omega'}$ is the original image $I$ subtracting clothing region $\mathcal{M}_c$. Note that the composited body mask $\mathcal{M}^{C}_{\omega}$ always keeps a similar layout with synthesized body part mask $\mathcal{M}_{\omega}^S$ by composition to eliminate the misaligned pixels in $\mathcal{M}_{\omega}^S$ .
It precisely preserves the non-target body part by combining the two masks (\ie $\mathcal{M}_{\omega}^S$ and $\mathcal{M}_{\omega}$), which are used to fully recover the non-targeted details in the following step to fully preserve $\mathcal{I}_\omega$ and generate coherent body parts with the guidance of $ \mathcal{M}^{G}_{a}$. It is also worth noting that it can adaptively deal with different cases. For example, when transferring a T-shirt (short-sleeve) to a person in long-sleeve only the within region of $\mathcal{M}^{G}_{a}$ will perform generation and preserve all the others, while in the opposite case, $\mathcal{M}^{G}_{a}=\mathbf{0}$ and surfeit body parts will be shaded by clothes as in Eq.~(\ref{euq:labelmask}) and Eq.~(\ref{euq:mask}).

\begin{figure}[htb]
\begin{center}
\vspace{-5pt}
\includegraphics[width=0.8\linewidth]{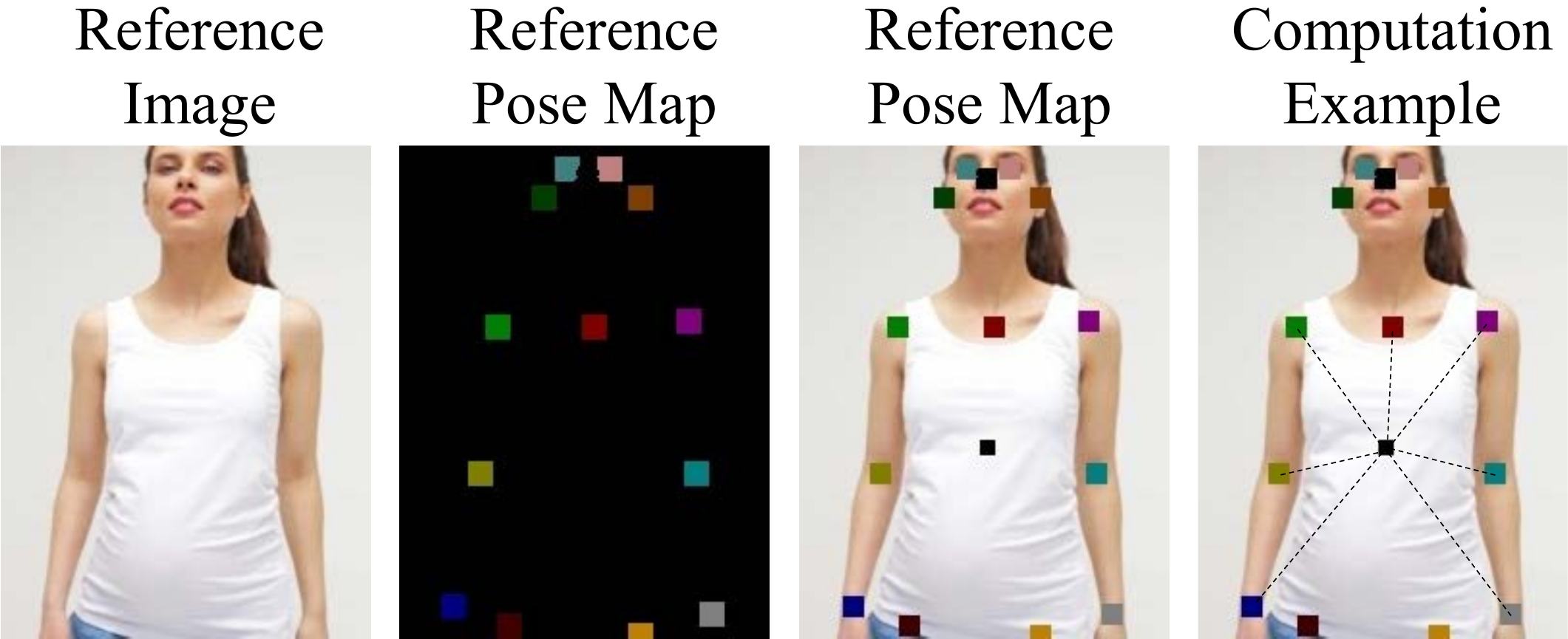}
\vspace{-10pt}
\end{center}
   \caption{\footnotesize An example of computing the complexity score $\mathcal{C}$. Given a reference image and its pose map, the connected points shown in the last image are selected to calculate $\mathcal{C}$ for the reference image.}
\label{fig:complexity}
\vspace{-10pt}
\end{figure}

\textbf{Mask Inpainting}. In order to fully exploit the layout adaptation ability of the network during training, CFM uses masks $\mathcal{M}_k$ from the \emph{Irregular Mask Dataset}~\cite{liu2018image} to randomly remove part of the arms in the body images $\mathcal{I}_{\omega}$ as $\mathcal{I}_{\omega}=(1-\mathcal{M}_k \odot \mathcal{M}_a)\odot \mathcal{I}_{\omega'}$  for mimicking image inpainting, where $\mathcal{M}_a$ is the mask of arms and this is similar to the Eq.~(\ref{euq:mask}) in the form, making it possible to separate the regions of preservation and generation. To combine the semantic information, composited body mask $\mathcal{M}^{C}_{\omega}$ and synthesized clothing mask $\mathcal{M}^{S}_{c}$ are concatenated with the body part image $\mathcal{I}_{\omega}$ and refined clothing  image $\mathcal{T}^{R}_{c}$ as the input.
Thus, the texture information can be recovered by the proposed inpainting based fusion GAN, yielding the photo-realistic results. Therefore,
in the inference stage, the network can adaptively generate the photo-realistic try-on image with rich details via the proposed CFM. Extensive experiments have proved that the proposed method can not only solve cases of easy and medium levels but hard cases with significant improvement.


\begin{figure*}[t]
\begin{center}
\includegraphics[width=1\linewidth]{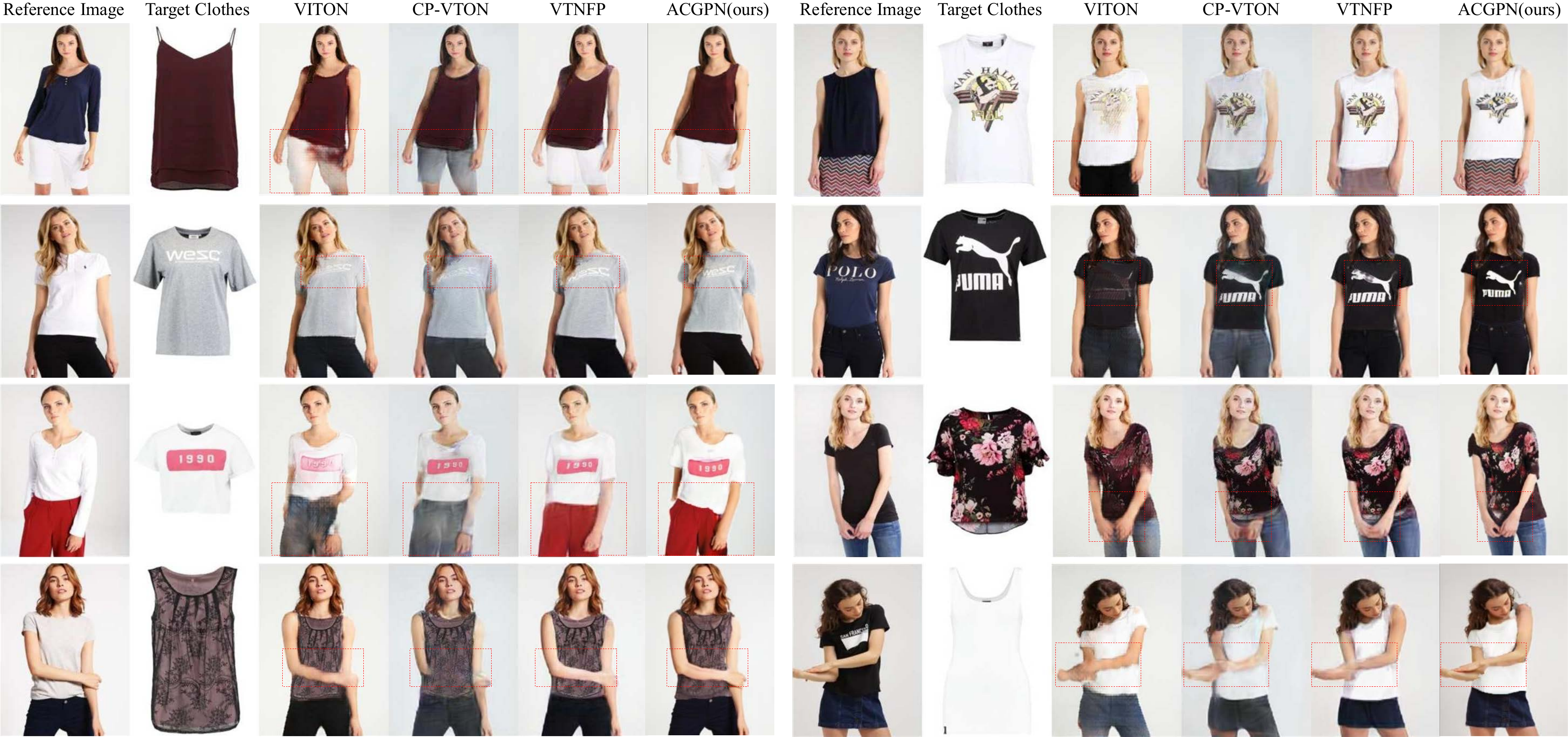}
\vspace{-10pt}
\end{center}
   \caption{\footnotesize Visual comparison of four virtual try-on methods on easy to hard levels (from top to bottom). ACGPN generates photo-realistic try-on results, which preserves both
the clothing texture and person body features. With the second-order difference constraint, the embroideries and texture are less likely to be distorted (\ie the 2nd row). With the preservation ability of the non-target body part composition, the body parts in our results are visually much more photo-realistic (\ie the 4th row). Especially different regions are marked in red-boxes. Note that the presented images are the same in the paper VTNFP~\cite{Yu_2019_ICCV} for fair comparison.}
\label{fig:Comparison}
\vspace{-10pt}
\end{figure*}

\section{Experiments}

\subsection{Dataset}
Experiments are conducted on the dataset (\ie VITON~\cite{DBLP:conf/cvpr/HanWWYD18} dataset) that used in VITON~\cite{DBLP:conf/cvpr/HanWWYD18} and CP-VITON~\cite{DBLP:conf/eccv/WangZLCLY18}. It contains about 19,000 image pairs, each of which includes a front-view woman image and a top clothing  image. After removing the invalid image pairs, it yields 16,253 pairs, further splitting into a training set of 14,221 pairs and a testing set of 2,032 pairs. ACGPN is compared with VITON, CP-VTON and VTNFP. Without the official code of VTNFP, we compare the visual results reported in VTNFP's paper and reproduce it for qualitative comparison. Extensive ACGPN try-on results are given in appendix.

\textbf{Dataset Partition}.
Images of try-on task exhibit different difficulty levels as shown in Fig.~\ref{fig:firstpage}. Easy case usually shows a standard posture with face forward and hands down; Medium level images present the twisting of the body torso or one of the hands overlapping with the body; and hard cases show both torso twisting and two hands blocking in front of the body. Limbs intersections and torso occlusions raise great challenge for the semantic layout prediction. To describe this, we propose to use reference points to represent body parts by leveraging pose maps as illustrated in Fig.~\ref{fig:complexity}. To quantitatively score the complexity of each image, we define the complexity of a certain image as
\vspace{-5pt}
\begin{equation}
\mathcal{C}=\frac{\sum_{t\in \mathcal{M}_{p'}}^N \left\lVert t-\frac{\sum_{t\in \mathcal{M}_{p'}}^Nt}{N}\right\rVert_1}{N},
\end{equation}
where $\mathcal{M}_{p'}$ represents points of left (right) arm, left (right) shoulder, left (right) hand, and torso. $t=\left(x_t,y_t\right)$ is a certain pose point and $N=7$ indicates the number of reference points. We define the thresholds of easy to medium as 80, and medium to hard as 68, in that when $\mathcal{C}<68$ the layout intersections become complicated, and when $\mathcal{C}>80$ the images tend to be standard posture, face forward and hands down. According to the statistics, 423, 514, 1095 images are split into hard, medium and easy level, respectively.
\vspace{-5pt}

\subsection{Implementation Details}

\textbf{Architecture}.~ACGPN contains SGM, CWM and CFM. All the generators in SGM and CFM have the same structure of U-Net~\cite{ronneberger2015u} and all the discriminators are from pix2pixHD~\cite{DBLP:conf/cvpr/Wang0ZTKC18}. The structure of STN~\cite{jaderberg2015spatial} in CWM begins with five convolution layers followed by a max-pooling layer with stride 2. Resolution for all images in training and testing is 256 $\times$ 192. Followed by steps in Fig.~\ref{fig:2}, we first predict the semantic layout of the reference image, and then decide the generation and preservation of image content. 

\textbf{Training}.
We train the proposed modules separately and combine them to eventually output the try-on image. Target clothes used in the training process are the same as in the reference image since it is intractable to have the ground-truth images of try-on results. Each module in the proposed method is trained for 20 epochs by setting the weights of losses $\lambda_{r}=\lambda_{s}=0.1$, $\lambda_{1}=\lambda_{2}=1$, and batch-size 8. The learning rate is initialized as 0.0002 and the network is optimized by Adam optimizer with the hyper-parameter $\beta_1 = 0.5$, and $\beta_2 = 0.999$. All the codes are implemented by deep learning toolkit PyTorch and eight NVIDIA 1080Ti GPUs are used in our experiments.

\textbf{Testing}.
The testing process follows the same procedure of training but is only different with that the target clothes are different from the ones in the reference images. We test our model in easy, medium and hard cases, respectively, and evaluate the results qualitatively and quantitatively. More evaluation results are given in the following sections.


\subsection{Qualitative Results}
We perform visual comparison of our proposed method with VITON~\cite{DBLP:conf/cvpr/HanWWYD18}, CP-VTON~\cite{DBLP:conf/eccv/WangZLCLY18}, and VTNFP~\cite{Yu_2019_ICCV}. As shown in Fig.~\ref{fig:Comparison}, from top to bottom, the difficulty levels of the try-on images are arranged from easy to hard.
In all difficulty levels the images generated by VITON show many visual artifacts including color mixture, boundary blurring, cluttered texture and so on. In comparison to VITON, CP-VITON achieves better visual results on easy level but still results in unnecessary editing on bottom clothes and blurring on body parts on medium and hard levels.
Bad cases such as broken arms in the generated images can also be observed when there are intersections between arms and torso. To sum up, VITON and CP-VTON warp the image onto the clothing region and map the texture and embroiders, thereby possibly causing the incorrect editing on body parts and bottom clothes.

VTNFP uses segmentation representation to further preserve the non-target details of body parts and bottom clothes, but is still limited to fully preserve the details, resulting in blurry output. The drawbacks behind VTNFP lie in the unawareness of the semantic layout and relationship within the layout, therefore being unable in extracting the specific region to preserve. In comparison to VITON and CP-VTON, VTNPF is better in preserving the character of clothes and visual results, but still struggles to generate body parts details (\ie hands and finger gaps).
It is worth noting that all the methods cannot avoid distortions and misalignments on the Logo or embroidery, remaining a large gap to photo-realistic try-on.

In contrast, ACGPN performs much better in simultaneously preserving the character of clothes and the body part information. Benefited from the proposed second-order spatial transformation constraint in CWM, it prevents Logo distortion and realizes character retention, making the warping process to be more stable to preserve texture and embroideries.
As shown in the first example of the second row in Fig.~\ref{fig:Comparison}, Logo ‘WESC’ is over-stretched in results of the competing methods; however, in ACGPN, it is clear and undistorted.
The proposed inpainting-based CFM specifies and preserves the unchanged body parts directly. Benefited from the prediction of semantic layout and adaptive preservation of body parts, ACGPN is able to preserve the fine-scale details which are easily lost in the competing methods, clearly demonstrating its superiority against VITON, CP-VTON and VTNFP.


\subsection{Quantitative Results}

We adopt Structural SIMilarity (SSIM) ~\cite{wang2004image} to measure the similaity between synthesized images and ground-truth, and Inception Score (IS)~\cite{DBLP:conf/nips/SalimansGZCRCC16} to measure the visual quality of synthesized images. Higher scores on both metrics indicate higher quality of the results.

Table ~\ref{tab:SSIM} lists the SSIM and IS scores by VITON~\cite{DBLP:conf/cvpr/HanWWYD18}, CP-VTON~\cite{DBLP:conf/eccv/WangZLCLY18}, VTNFP~\cite{Yu_2019_ICCV}, and our ACGPN.
Unsurprisingly, the SSIM score decreases along with the increase of difficult level, demonstrating the negative correlation between difficulty level and try-on image quality. Nonetheless, our ACGPN outperforms the competing methods by a large margin on both metrics for all difficulty levels.
For the easy case, ACGPN surpasses VITON, CP-VTON and VTNFP by 0.067, 0.101 and 0.044 in terms of SSIM. For the medium case, the gains by ACGPN are 0.062, 0.099 and 0.040.
As for the hard case, ACGPN also outperforms VITON, CP-VTON and VTNFP by 0.049, 0.099 and 0.040.
In terms of IS, the overall gains against VITON, CP-VTON  and VTNFP are respectively 0.179, 0.072 and 0.045, further showing the superiority of ACGPN by quantitative metrics.


\begin{table}
\renewcommand\tabcolsep{3.0pt} 
\footnotesize

\begin{center}
\begin{tabular}{lccccc}
\hline
\multirow{2}*{Method}&\multicolumn{4}{c}{SSIM}&\multirow{2}{*}{IS}\\
\cline{2-5}
 & All &Easy&Medium&Hard \\

\hline\hline

VITON~\cite{DBLP:conf/cvpr/HanWWYD18}&0.783&0.787&0.779&0.779&2.650
\\
CP-VTON~\cite{DBLP:conf/eccv/WangZLCLY18}&0.745&0.753&0.742&0.729&2.757\\
VTNFP~\cite{Yu_2019_ICCV}&0.803&0.810&0.801&0.788&2.784\\
\hline
ACGPN$\dag$ &0.825&0.834&0.823&0.805&2.805\\
					
ACGPN*&0.826&0.835&0.823	&0.806&2.798\\
ACGPN&\textbf{0.845}&\textbf{0.854}&\textbf{0.841}&\textbf{0.828}&\textbf{2.829}\\
\hline
\end{tabular}
\caption{\footnotesize The SSIM~\cite{wang2004image} and IS~\cite{DBLP:conf/nips/SalimansGZCRCC16} results of five methods. ACGPN$\dag$ and ACGPN* are ACGPN variants for ablation study. }
\label{tab:SSIM}
\vspace{-25pt}
\end{center}
\end{table}


\subsection{Ablation Study}

\begin{figure}[htb]
\begin{center}
\vspace{-5pt}
\includegraphics[width=0.9\linewidth]{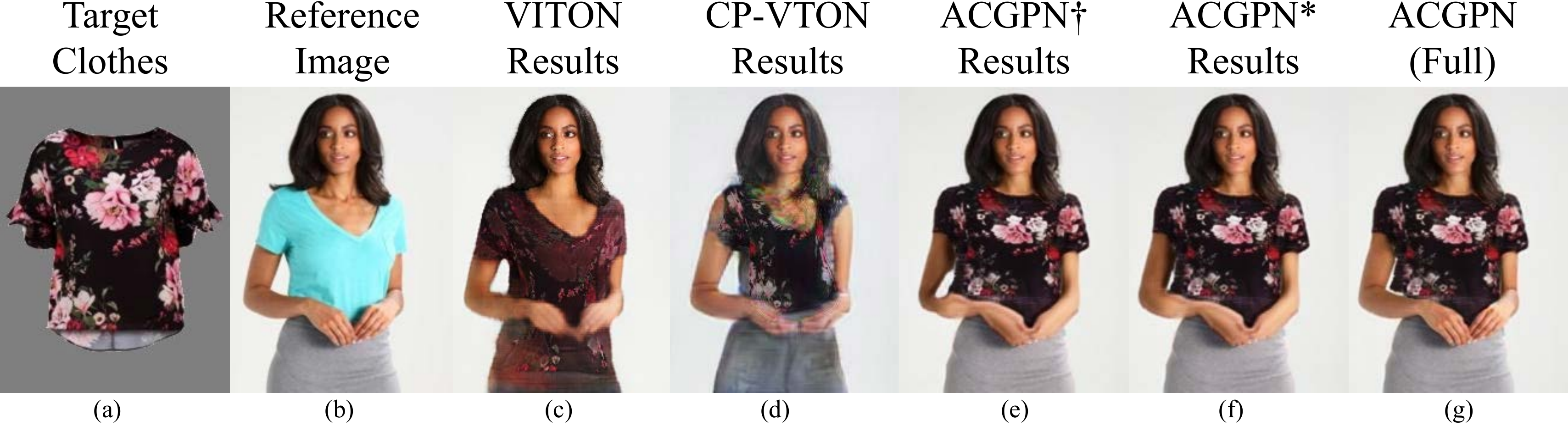}
\vspace{-10pt}
\end{center}
   \caption{\footnotesize Visual comparison of our non-target body part composition. (c) generates incorrect target clothes and blurry body parts; (d) produces body parts with deformation; (e) and (f) show some distorted body parts; (g) generates the convincing result.}
\label{fig:preservation}
\vspace{-10pt}
\end{figure}

Ablation study is conducted to evaluate the effectiveness of the major modules in ACGPN in Table ~\ref{tab:SSIM}.
Here, ACGPN$\dag$ refers to directly using $\mathcal{M}^{S}_{\omega}$ instead of $\mathcal{M}^{C}_{\omega}$ in CFM to generate try-on image, and ACGPN* refers to using $\mathcal{M}^{C}_{\omega}$ as the input. Both models use $\mathcal{I}_{\omega}$ with the removal of arms. Comparing to ACGPN$\dag$, ACGPN* and ACGPN, it shows that the non-target body part composition indeed contributes to better visual results.
We also notice that ACGPN$\dag$ and ACGPN* also outperform VITON~\cite{DBLP:conf/cvpr/HanWWYD18} and CP-VTON~\cite{DBLP:conf/eccv/WangZLCLY18} by a margin, owing to the accurate estimation of semantic layout.
Visual comparison results in Fig.~\ref{fig:preservation} further show the effectiveness of body part composition in adaptive preservation.
With the composition, the human body layout can be clearly stratified. Otherwise, we can only get correct body part shape but may generate wrong details as in (f) of Fig.~\ref{fig:preservation}.

\begin{figure}[htb]
\begin{center}
\includegraphics[width=0.9\linewidth]{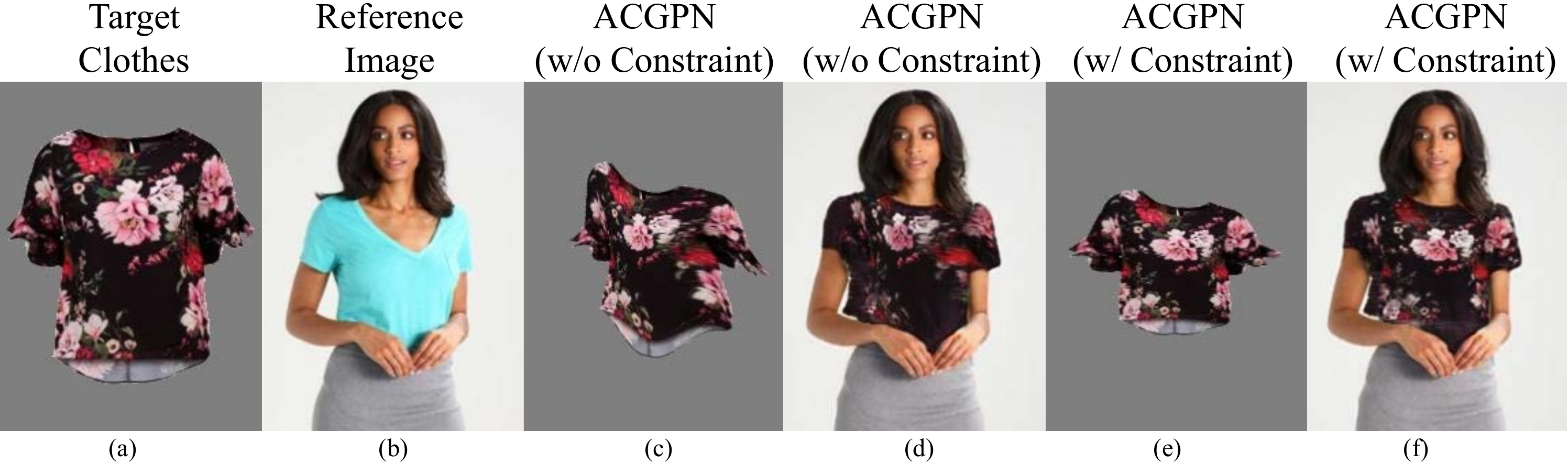}
\vspace{-10pt}
\end{center}
   \caption{\footnotesize Ablation study on the effect of second-order difference constraint. (c), (e) are the warped clothes, and (d), (f) are the synthesized results. Although ACGPN eliminates the artifacts in distorted warped clothing  image (c), it still largely influences its verisimilitude of (d). }
\label{fig:AblationWarp}
\vspace{-10pt}

\end{figure}

Experiment is also conducted to verify the effectiveness of our second-order difference constraint in CWM. As shown in Fig.~\ref{fig:AblationWarp}, we choose target clothes with complicated embroiders as examples.
From Fig.~\ref{fig:AblationWarp}(c), the warping model may generate distorted images without the constraint.

It is worth noting that, due to the effectiveness of semantic layout prediction, ACGPN without the constraint can still produce satisfying result, and the target clothes with pure color or simple embroideries are less vulnerable to the degeneration of warping.
As for the target clothes with complex textures, the second-order difference constraint plays an important role in generating photo-realistic results with correct detailed textures
(see in Fig.~\ref{fig:AblationWarp}(d)(f)).

\begin{table}
\renewcommand\tabcolsep{4pt}
\begin{center}
\footnotesize
\begin{tabular}{lcccc}
\hline
Method&Easy&Medium&Hard&Mean\\
\hline\hline
CP-VTON~\cite{DBLP:conf/eccv/WangZLCLY18}&15.4\%&11.2\%&4.0\%&10.2\%\\
ACGPN&\textbf{84.6\%}&\textbf{88.8\%}&\textbf{96.0\%}&\textbf{89.8\%}\\
\hline
VITON~\cite{DBLP:conf/cvpr/HanWWYD18}&38.8\%&18.2\%&13.3\%&23.4\%\\
ACGPN&\textbf{61.2\%}&\textbf{81.8\%}&\textbf{86.7\%}&\textbf{76.6\%}\\
\hline
VTNFP~\cite{Yu_2019_ICCV}&45.6\%&31.0\%&23.4\%&33.3\%\\
ACGPN&\textbf{54.4\%}&\textbf{69.0\%}&\textbf{76.6\%}&\textbf{66.7\%}\\

\hline

\end{tabular}

\caption{\footnotesize User study results on the VITON dataset. The percentage indicates the ratio of images which are voted to be better than the compared method.}
\label{tab:User_study}
\vspace{-20pt}
\end{center}
\end{table}

\subsection{User Study}
To further assess the results of try-on images generated by VITON~\cite{DBLP:conf/cvpr/HanWWYD18}, CP-VTON~\cite{DBLP:conf/eccv/WangZLCLY18}, and ACGPN, we conduct a user study by recruiting 50 volunteers. We first test 200 images by different methods from easy, medium, and hard cases, respectively, and then group 1,800 pairs in total (each method contain 600 test images for three levels and each pair includes images from different methods).
Each volunteer is assigned 100 image pairs in an A/B manner randomly.
For each image pair, the target clothes and reference images are also attached in the user study. Each volunteer is asked to choose a better image meeting three criterion : (a) how well the target clothing character and posture of reference image are preserved; (b) how photo-realistic the whole image is; (c) how good the whole person seems. And we give the user unlimited time to choose the one with better quality. The results are shown in Table ~\ref{tab:User_study}. It reveals the great superiority of ACGPN over the other methods, especially on hard cases. The results demonstrate the effectiveness of the proposed method in handling body part intersections and occlusions on visual try-on tasks.

\section{Conclusion}
In this work, we propose a novel adaptive content generating and preserving Network, \ie ACGPN, which aims at generating photo-realistic try-on result while preserving both the character of clothes and details of human identity (posture, body parts, bottom clothes).
We present three carefully designed modules, \ie Mask Generation Module (GMM), Clothes Warping Module (CWM), and Content Fusion Module (CFM).
We evaluate our ACGPN on the VITON~\cite{DBLP:conf/cvpr/HanWWYD18} dataset with three levels of try-on difficulty.
The results clearly show the great superiority of ACGPN over the state-of-the-art methods in terms of quantitative metrics, visual quality and user study.

\small{
\textbf{Acknowledgement}
This work was partially supported by HKU Seed Fund for Basic Research and Start-up Fund, and the NSFC project under Grant No. U19A2073. }

{\small
\bibliographystyle{ieee_fullname}
\bibliography{ms}
}


\begin{figure}[htb]

\twocolumn[{
\renewcommand\twocolumn[1][]{#1}%
\vspace{-30pt}

\begin{center}
  \centering
  \includegraphics[width=1\textwidth]{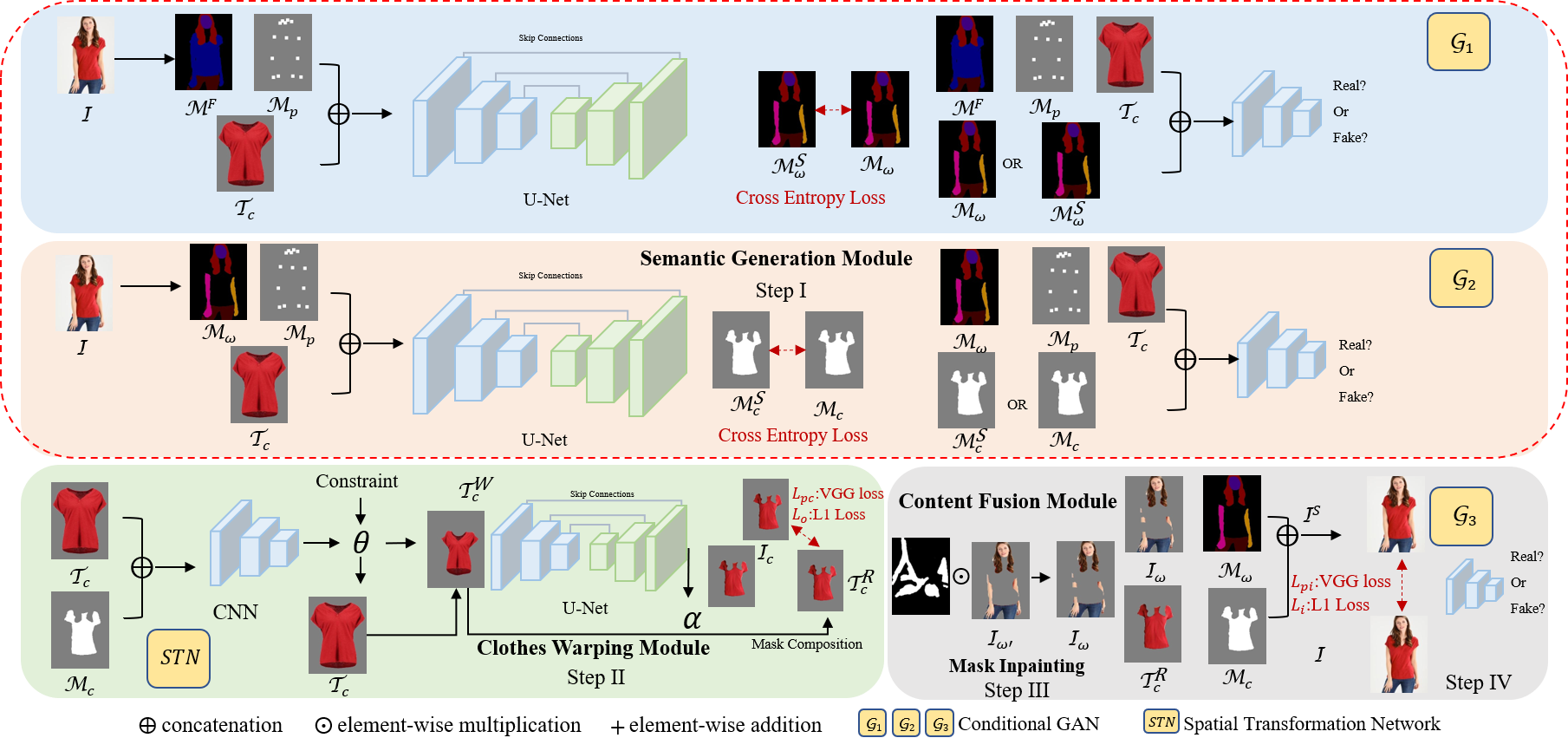}
  \vspace{-20pt}

\end{center}

\caption{\footnotesize The overall training pipeline of ACGPN. Two major differences from the inference are that, first, the target clothes $\mathcal{T}_c$ is the in-shop version of the clothes on reference image $\mathcal{I}$, second, the Step III masks the limbs of $\mathcal{I}_{\omega'}$ (Reference person $\mathcal{I}$ removing clothes) to form $\mathcal{I}_{\omega}$ which is later fed into Step IV as input. The reconstruction loss for semantic masks and RGB images (\ie clothing images, clothed person images) are cross entropy loss~\cite{goodfellow2016deep} and perceptual loss~\cite{JohnsonPerceptual} (\ie VGG loss) combined with L1 loss (pixel-wise $\mathcal{L}_1$ distance).
}
\label{fig:pipeline}
  \vspace{20pt}
}]
\end{figure}
\vspace{-20pt}

\thispagestyle{empty}

\section{The Reconstruction loss}
We can see the whole training pipeline in Fig.~\ref{fig:pipeline} which includes all the losses, the interactions between generator and discriminator and the compositions of inputs as well as outputs. Here we introduce the reconstruction losses for training in Step II and Step IV, which are widely used in most of the image-to-image translation tasks.

In step II shown in Fig.~\ref{fig:pipeline}, in order to refine the $\mathcal{T}_c^W$, a U-Net is used to refine the warped clothes to fit the mask $\mathcal{M}_c$. $\mathcal{T}_c^W$ and $\mathcal{M}_c$ are fed into the refinement network and a coarse result as well as a composition mask $\alpha$ will be produced to perform the composition,
\begin{equation}
    \mathcal{T}^{R}_{c}=(1-\alpha)\odot\mathcal{T}^{W}_{c}+\alpha \odot\mathcal{T}^{R}_{c},
\label{refineclothes}
\end{equation}
where $\odot$ indicates element-wise multiplication.

\begin{figure*}[htb]
\begin{center}
\includegraphics[width=1\linewidth]{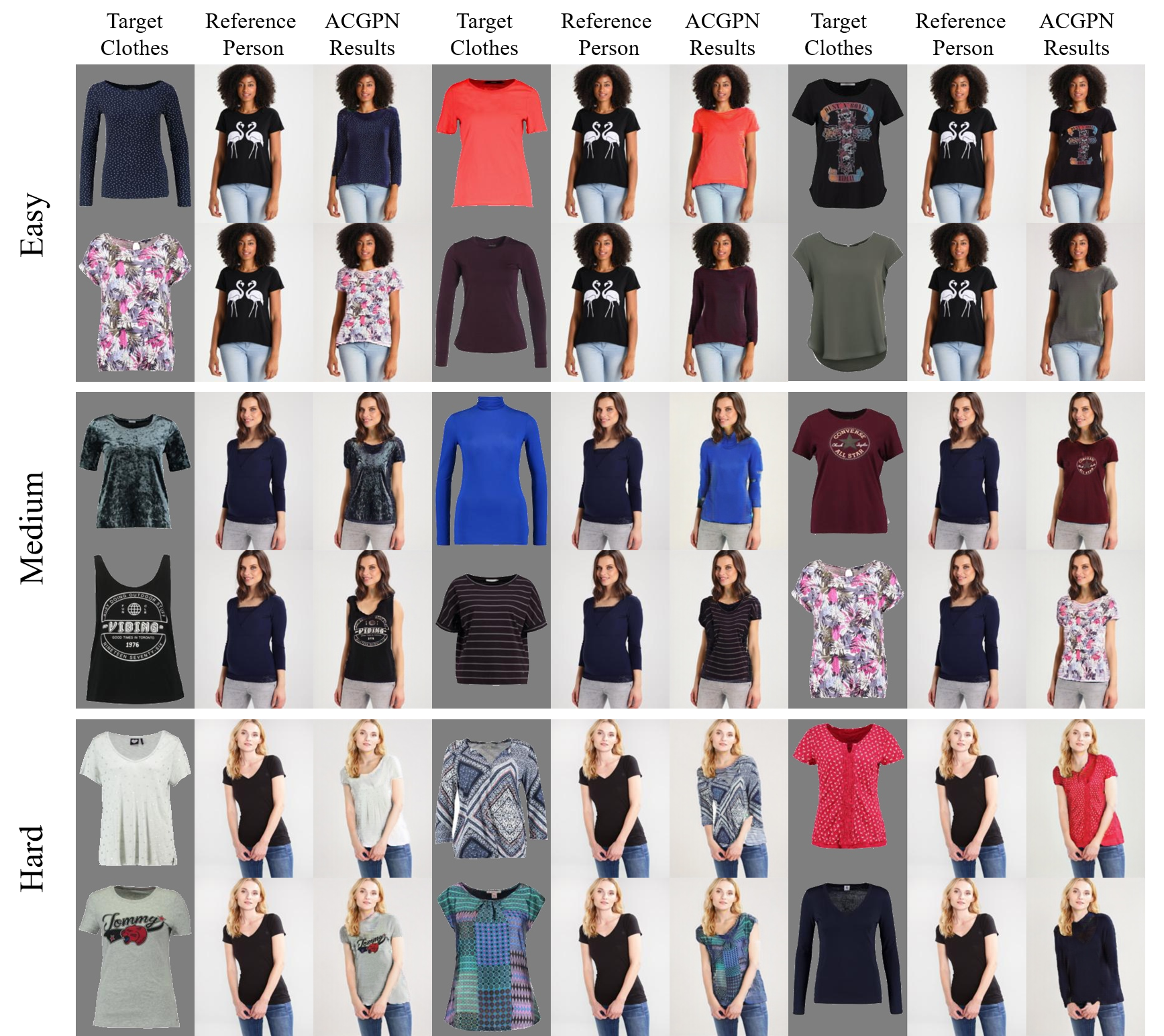}
\end{center}
   \caption{Extensive try-on results with three difficulty levels. We can see that ACGPN performs equally well for long-sleeve clothes to short-sleeve reference image (fifth row in the middle) and short-sleeve clothes to long-sleeve reference image (fourth row on the left), which demonstrates the generality of our method.
}
\label{fig:more}
\end{figure*}

The loss for the refinement operation is the combination of $L_1$ loss and perceptual loss~\cite{JohnsonPerceptual} (\ie VGG loss which computes the distance of the features extracted by VGG19~\cite{Simonyan2014Very}).
\begin{equation}
    \mathcal{L}_{o}=\lVert \mathcal{T}^{R}_{c} \!-\! \mathcal{I}_c \rVert_1,
\end{equation}
where $\mathcal{L}_{o}$ indicates the $L_1$ loss in refinement of $\mathcal{T}^{W}_{c}$, and $\mathcal{I}_c$ is the ground-truth. And the full reconstruction loss to refine $\mathcal{T}^{W}_{c}$is
\begin{equation}
    \mathcal{L}_{rc}=\lambda_o\mathcal{L}_o+\lambda_{pc}\mathcal{L}_{pc},
\end{equation}
where $\mathcal{L} _{rc}$ indicates the reconstruction loss of $\mathcal{T}^{R}_{c}$, and $\mathcal{L}_{pc}$ is the perceptual loss between $\mathcal{T}^{R}_{c}$ and $\mathcal{I}_c$. $\lambda_{o}$ and $\lambda_{pc}$ are weights of each loss.

In Step IV shown in Fig.~\ref{fig:pipeline}, the reconstruction loss for the inpainting based fusion GAN is also the combination of $L_1$ loss and perceptual loss between the synthesized image and the ground-truth image. The $L_1$ loss $\mathcal{L}_i$ is formulated as
\begin{equation}
    \mathcal{L}_i=\lVert \mathcal{I}^S\!-\!\mathcal{I} \rVert_1.
\end{equation}
The full reconstruction loss is
\begin{equation}
     \mathcal{L}_{ri}= \lambda_{i}\mathcal{L}_i+\lambda_{pi} \mathcal{L}_{pi},
\end{equation}
where $\mathcal{L}_{pi}$ is the perceptual loss between $\mathcal{I}^S$ and its ground-truth $\mathcal{I}$, and $ \mathcal{L}_{ri}$ indicates the full reconstruction loss of $\mathcal{I}^S$.  $\lambda_{i}$ and $\lambda_{pi}$ are weights of each loss.

The weights of each loss are given as $\mathcal{L}_{o}\!=\!\mathcal{L}_{i}\!=\!1$ and $\mathcal{L}_{pc}\!=\!\mathcal{L}_{pi}\!=\!10$.
\begin{figure*}[htb]
\begin{center}
\includegraphics[width=0.9\linewidth]{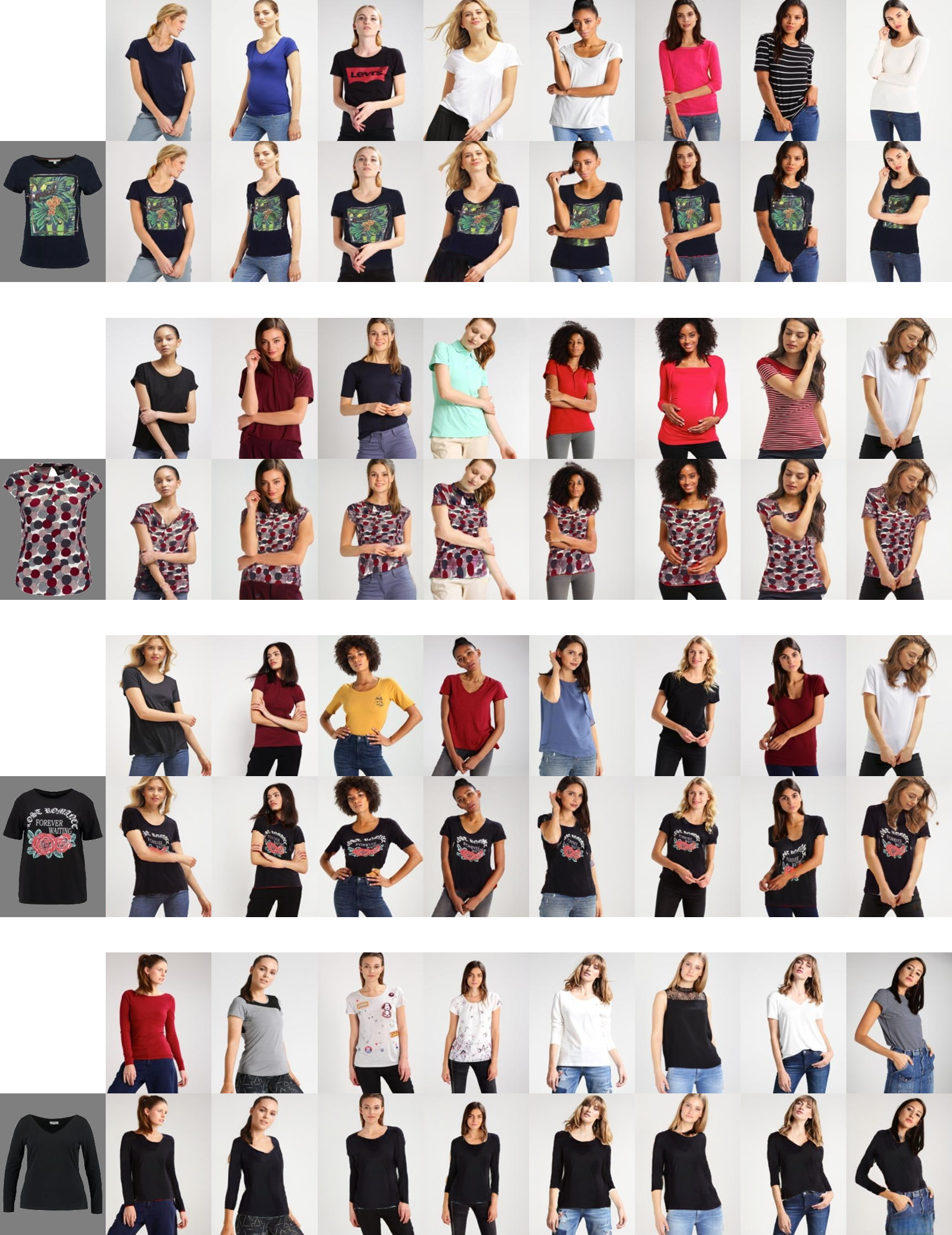}
\end{center}
   \caption{Extensive try-on results with four reference people. ACGPN perform robustly with various poses including occlusions and cross-arms. Artifacts are reduced to the minimum.
}
\label{fig:exten}
\end{figure*}
\section{More Try-on Results}
We here show more try-on results produced by ACGPN in Fig.~\ref{fig:more} and Fig.~\ref{fig:exten}. For more results, \textbf{an example video is provided in youtube}:
\url{https://www.youtube.com/watch?v=h-QWM92VLA0}.


\end{document}